\documentclass[10pt,twocolumn]{article}
\usepackage{arxiv}
\usepackage{times}  
\usepackage{helvet}  
\usepackage{courier}  
\usepackage[hyphens]{url}  
\usepackage{graphicx} 
\usepackage{natbib}  
\usepackage{caption} 
\DeclareCaptionStyle{ruled}{labelfont=normalfont,labelsep=colon,strut=off} 
\title{Breaking the Softmax Bottleneck for Session-based Recommender Systems with Dropout and Decoupling}
\author {
  Ying-Chen Lin \\
  Independent Researcher
}

\usepackage{booktabs}
\usepackage{multirow}
\usepackage{amsmath}
\usepackage{amsfonts}
\usepackage{subcaption}
\usepackage{nicefrac}

\newcommand\B[1]{\textbf{#1}}

\newcommand\U[1]{\underline{#1}}

\usepackage{tikz}
\usetikzlibrary{positioning, calc}
\tikzstyle{block}=[draw, rounded corners, thick, fill=black!5, shape=rectangle, font=\sffamily]
\tikzstyle{label}=[font=\sffamily]

\begin{document}

\maketitle

\begin{abstract}
The Softmax bottleneck was first identified in language modeling as a theoretical limit on the expressivity of Softmax-based models. Being one of the most widely-used methods to output probability, Softmax-based models have found a wide range of applications, including session-based recommender systems (SBRSs). Softmax-based models consist of a Softmax function on top of a final linear layer. The bottleneck has been shown to be caused by rank deficiency in the final linear layer due to its connection with matrix factorization.
In this paper, we show that there’s more aspects to the Softmax bottleneck in SBRSs. Contrary to common beliefs, overfitting does happen in the final linear layer, while it’s often associated with complex networks. Furthermore, we identified that the common technique of sharing item embeddings among session sequences and the candidate pool creates a tight-coupling that also contributes to the bottleneck.
We propose a simple yet effective method, Dropout and Decoupling (D\&D), to alleviate these problems. Our experiments show that our method significantly improves the accuracy of a variety of Softmax-based SBRS algorithms. When compared to other computationally expensive methods, such as MLP and MoS (Mixture of Softmaxes), our method performs on par with and at times even better than those methods, while keeping the same time complexity as Softmax-based models.
\end{abstract}

\section{Introduction}

Recommender systems are essential tools for online services to guide users through vast volumes of digital contents. As technology enables the Internet to become faster and more ubiquitous, the amount of online data is growing at an exponential rate. This rapid growth imposes challenges to both online service providers and their users with information overload. Designed to serve the purpose of filtering and prioritizing useful information, recommender systems have since found a wide range of applications in social media, eCommerce, and streaming services.

The Softmax bottleneck is a theoretical limit on the expressivity of Softmax-based models. It was first discovered by \citeauthor{softmax} in language modeling. Language modeling faces a similar challenge as recommender systems when selecting the next word from a massive vocabulary. A linear-Softmax output layer is often adopted in these classification problems, when the number of labels is several orders of magnitudes larger than the dimension of hidden states. It consists of a Softmax function around logits, which are dot products between the hidden state and candidate word embeddings. The authors pointed out that its linearity causes the learned logit matrix to be low-rank. As a result, it limits the expressivity of the whole network, hence getting the name of the Softmax bottleneck.

Linear-Softmax layers are widely used in session-based recommender systems (SBRSs), which take a sequence of items in the session as input and predict the next item from a candidate pool. There are several reasons behind this. 
First, the accuracy of SBRSs really benefits from learning from the whole candidate pool, instead of having a candidate selection mechanism. Since the candidate pool is often massive in size, a linear-Softmax layer is almost the only computationally efficient choice.
Second, the simplicity of linear-Softmax layers allows them to be accelerated substantially in production. The monotonic property of the Softmax function allows the logits to be used directly for ranking, and modern search engines support dot product ranking through spatial indices with logarithmic time complexity.
The above reasons describe why Softmax output layers are almost unanimously adopted in SBRS research.

In this paper, we introduce more aspects of the Softmax bottleneck other than expressivity in the context of SBRSs.
First, overfitting does occur in linear-Softmax layers. Often associated with complex and flexible networks, we show that overfitting could be a serious issue even in a dot product. We also show that with proper dropout layers, the accuracy of the whole network could be improved significantly on a variety of SBRSs.
Second, although not always stated explicitly, the technique of sharing the item embeddings among session sequences and the candidate pool is almost adopted universally in research \cite{narm, stamp, srgnn} to boost performance. This tight-coupling simultaneously expects the embeddings to be the encoder inputs and also the weights in dot products. It could create interference that stops accuracy from further improvement. We show that a single layer of feedforward network could alleviate this problem.

Combining the above two ideas, we propose Dropout and Decoupling (D\&D), a simple yet effective method that  alleviates the previously mentioned problems. It also keeps the time complexity of linear-Softmax output layers in both training and query time. We compare our method to other computationally expensive methods, including MLP and MoS from \cite{softmax}, and show that D\&D performs on par with and at times even better than those methods.
Our main contributions are summarized as follows:
\begin{itemize}
  \item We identify that overfitting is a prominent problem in Softmax-based session-based recommender systems.
  \item We show that sharing the item embedding layer among session sequences and the candidate pool introduces a tight-coupling that restrains performance in session-based recommender systems.
  \item We propose Dropout and Decoupling to effectively alleviate the above problems while keeping the same time complexity during both training and query.
\end{itemize}

\section{Related Work}

In this section, we will review related works on the Softmax bottleneck and recommender systems.

\subsection{The Softmax Bottleneck}

The Softmax bottleneck was first described by the work \cite{softmax} in language modeling. The authors identified linear-Softmax output layers as the source of limited expressivity. Natural languages are commonly believed to be highly diverse and context dependent \cite{rnnml}. Modeling the probability logits using matrix factorization causes rank deficiency. To tackle this problem, MoS (mixture of Softmaxes) was proposed. As weighted mixtures of multiple Softmax components, MoS improves on both perplexity and rank of logit matrices. Several later works \cite{mixtape, sigsoftmax, lmsplif} have proposed lightweight alternatives to the computationally expensive MoS, and also techniques to further improve MoS.

\subsection{Session-based Recommender Systems}

Session-based recommender systems learn user behavior from sequence data. Pioneer works in this field employs statistical models to capture sequential patterns. The problem has been formulated as decision tree learning \cite{mc} and Markov decision process (MDP) \cite{mdp}. Later works \cite{fpmc, fossil} combine Markov chain with matrix factorization (MF) to model personalized behavior patterns.

Deep learning provides more sophisticated methods to encode user behavior sequences. RNN-based approaches \cite{gru4rec} were introduced to model sequential behavior. Later work \cite{gru4recp} enhances performance by proper data augmentation and considering temporal shift in data. CNN-based approaches \cite{caser} were also proposed. Dot-product attention \cite{narm, stamp} was introduced to capture user's current interests. Methods with self-attention \cite{sasrec, bert4rec} were proposed to capture complex sequence patterns. GNN-based approaches \cite{srgnn} were also introduced to capture complex transitions between items.

\subsection{Collaborative Filtering and CTR Prediction}

Collaborative filtering (CF) tasks model interactions between users and items. They predict future interactions with a given set of past interactions. Pioneer works \cite{pmf, bprmf} employed matrix factorization (MF) on user-item matrices. Autoencoder-based approaches \cite{cdae, vaecf} were introduced to produce more robust reconstruction of user interactions. NCF \cite{ncf} generalizes MF and uses it in combination with MLP.

CTR prediction tasks model click-through rate from more rich features. The features expand the data dimension and sparsity becomes a problem. To tackle this problem, wide \& deep network \cite{deepwide} was proposed to use MLP with cross-product features. Many later works were inspired by this method. DeepFM \cite{deepfm} combines MLP with a factorization machine (FM). DCN \cite{dcn} combines MLP with a proposed cross network.

\section{Softmax Bottleneck in Language Modeling}

We will briefly describe the Softmax bottleneck problem in language modeling, and the theoretical limit it imposes on the expressivity of the whole model.

\subsection{Language Modeling}

A language model consists of a vocabulary of words $\mathcal{V} = \left\{x_1, ..., x_M\right\}$ and a probability distribution $P$ defined over all sentences in $\mathcal{V}$, such as $\left(x_{i_1}, ..., x_{i_T}\right)$. With factorization, $P$ could be rewritten as a product of conditional distributions $P\left(X_1, ..., X_T\right) \prod_{t=1}^T P\left(X_t | C_t\right)$, where $C_t = X_{<t}$ is referred to as the context. This formula describes a random process which generates the next word in a given context. All sentences in the given corpus are assumed to be generated from a ground-truth distribution $P^\ast\left(X_t | C_t\right)$. The goal of language modeling is to approximate $P^\ast$ with a parametric distribution $P_\theta\left(X_t | C_t\right)$.

The majority of language models adopt a Softmax output layer \cite{softmax}. In these models, a context, or an input sequence, is represented by a $D$-dimensional vector $\mathbf{h}_j = f\left(c_j\right)$, where $c_j = \left(x_{j_1}, ..., x_{j_T}\right)$, and each word $x_i$ in the vocabulary are also represented by a $D$-dimensional vector $\mathbf{w}_i$. Then, the probability of the next word in a context is given by the \emph{linear-Softmax} formula in equation \ref{eq-lmprob}.

\begin{equation}
  P_\theta\left(x_i | c_j\right) = \frac
  {\exp\left(\mathbf{h}_j^\top \mathbf{w}_i\right)}
  {\sum^M_{i'=1} \exp\left(\mathbf{h}_j^\top \mathbf{w}_{i'}\right)}
  \label{eq-lmprob}
\end{equation}

\noindent $\theta$ is the parameters of the model. Training is done by optimizing the cross-entropy loss function in equation \ref{eq-lmloss}.

\begin{equation}
  L\left(\theta\right) = \frac{1}{N} \sum^N_{j=1} 
  -\log\left(Q_\theta\left(x_{i_j} | c_j\right)\right)
  \label{eq-lmloss}
\end{equation}

\subsection{The Softmax Bottleneck}

The Softmax bottleneck comes down to the assumption that the ground-truth distribution of a language model could be approximated by the dot product of two low-rank matrices passed through a Softmax function. 
Let $\mathbf{A} \in \mathbb{R}^{N\times M}$, $\mathbf{H} \in \mathbb{R}^{N\times D}$, $\mathbf{W} \in \mathbb{R}^{M\times D}$ denote the log of the ground-truth distribution, and the context representation matrix, the word embedding matrix respectively. In other words, $\mathbf{A}_{ji} = \log\left(P^\ast\left(x_i|c_j\right)\right)$, $\mathbf{H}_j = \mathbf{h}_j$, $\mathbf{W}_i = \mathbf{w}_i$. Let $F(\mathbf{A})$ denote the set of all matrices obtained by applying row-wise shift on $\mathbf{A}$. They will all produce the same distribution $P^\ast$ due to the normalization property of the Softmax function. The linear-Softmax formula can approximate $P^\ast$ if and only if $\mathbf{H}\mathbf{W}^\top$ approximates any matrix in $F(\mathbf{A})$.

As shown in the original work \cite{softmax}, the rank of $\mathbf{H}\mathbf{W}^\top$ could not be greater than the number of dimension $D$, but the rank of $\mathbf{A}$ could be any number up to $\min(N, M)$, which is usually several orders of magnitude larger than $D$. Combining with the fact $\left|\mathrm{rank}(\mathbf{A}') - \mathrm{rank}(\mathbf{A})\right| \le 1$ for all $\mathbf{A}' \in F(\mathbf{A})$, it's almost certainly impossible for $\mathbf{H}\mathbf{W}^\top$ to be in $F(\mathbf{A})$. This expressivity limit caused by this rank deficiency is called the Softmax bottleneck.

\subsection{Mixture of Softmaxes}

To overcome rank deficiency, \citeauthor{softmax} proposed Mixture of Softmaxes, or MoS, to improve expressivity. MoS is formulated as the weighted average over $K$ Softmax components, shown in equation \ref{eq-mosprob}. The authors have shown empirically that MoS could produce a high-rank logit matrix.

\begin{equation}
  P_\theta\left(x_i | c_j\right) = \sum_{k=1}^K \pi_{j,k} \frac
  {\exp\left(\mathbf{h}_{j, k}^\top \mathbf{w}_i\right)}
  {\sum_{i'=1}^M \exp\left(\mathbf{h}_{j, k}^\top \mathbf{w}_{i'}\right)}
  \label{eq-mosprob}
\end{equation}
where $\pi_{j,k}$ is the mixture weight of the $k$-th component and $\mathbf{h}_{j, k}$ is the $k$-th context vector, shown in equations \ref{eq-mosprob2}. The word embeddings are shared among all Softmax components to not significantly increase the model size.

\begin{equation}
  \begin{aligned}
    \pi_{j,k} &= \frac
    {\exp\left(\mathbf{h}_j^\top \mathbf{w}^{(\pi)}_k\right)}
    {\sum_{k'=1}^K \exp\left(\mathbf{h}_j^\top \mathbf{w}^{(\pi)}_{k'}\right)} \\
    \mathbf{h}_{j, k} &= \tanh\left(\mathbf{W}^{(h)}_k \mathbf{h}_j\right) \\
  \end{aligned}
  \label{eq-mosprob2}
\end{equation}

\section{Dropout and Decoupling} \label{sec-method}

We will identify more aspects of the Softmax bottleneck problem, besides just expressivity, that restrain the accuracy of session-based recommender systems. Then, we will introduce Dropout and Decoupling, a simple yet effective method, to alleviate those problems.

\subsection{Session-based Recommender Systems}

\begin{figure}[t]
  \centering
  \subfloat[Generic] {
    \begin{tikzpicture}
      \node (ses) at (-0.6, -0.2) [label] {\tiny Session};
      \node (ses) at (1.25, -0.2) [label] {\tiny Candidates};
      \node (emb) at (0, 0.8) [block, minimum width=3.0cm] {\tiny Shared Item Embedding};
      \node (sdo) at (-0.5, 1.4) [block, fill=black!15, minimum width=2.0cm] {\tiny Session Dropout};
      \node (edo) at (-0.5, 2.8) [block, fill=black!15] {\tiny Encoder Dropout};
      \node (enc) at (-0.5, 2.2) [block, minimum width=2.0cm] {\tiny Sequence Encoder};
      \node (dec) at (0, 3.5) [block] {\tiny Prob. Decocder};

      \draw [black!15, thick, fill=white] (1.05, 0.15) circle (0.15);
      \draw [black!15, thick, -latex] (1.05, 0.3) -- ($(emb.south)+(1.05, 0)$);
      \draw [black!30, thick, fill=white] (1.15, 0.15) circle (0.15);
      \draw [black!30, thick, -latex] (1.15, 0.3) -- ($(emb.south)+(1.15, 0)$);
      \foreach \x in {-1.25, -0.75, -0.25, 0.25, 1.25}{
        \draw [black, thick, fill=white] (\x, 0.15) circle (0.15);
        \draw [black, thick, -latex] (\x, 0.3) -- ($(emb.south)+(\x, 0)$);
      }
      \foreach \x in {-1.25, -0.75, -0.25, 0.25}{
        \draw [black, thick] ($(emb.north)+(\x, 0)$) -- ($(sdo.south)+(\x+0.5, 0)$);
        \draw [black, thick, -latex] ($(sdo.north)+(\x+0.5, 0)$) -- ($(enc.south)+(\x+0.5, 0)$);
      }
      \draw [black, thick] (enc.north) |- (edo.south);
      \draw [black, thick, -latex] (edo.north) -- ($(edo.north)+(0, 0.15)$) -- ($(edo.north)+(-0.75, 0.15)$) |- (dec.west);
      \draw [black, thick, -latex] ($(emb.north)+(1.25, 0)$) |- (dec.east);
    \end{tikzpicture}
    \label{fig-lin}
  }
  \quad
  \subfloat[Dropout and Decoupling] {
    \begin{tikzpicture}
      \node (ses) at (-0.6, -0.2) [label] {\tiny Session};
      \node (ses) at (1.25, -0.2) [label] {\tiny Candidates};
      \node (emb) at (0, 0.8) [block, minimum width=3.0cm] {\tiny Shared Item Embedding};
      \node (sdo) at (-0.5, 1.4) [block, fill=black!15, minimum width=2.0cm] {\tiny Session Dropout};
      \node (cdo) at (1.25, 1.6) [block, fill=black!15, align=center] {\tiny Candidate \\[-0.5ex] \tiny Dropout};
      \node (ffd) at (1.25, 2.8) [block, align=center] {\tiny Decoupling \\[-0.5ex] \tiny Feedforward*};
      \node (edo) at (-0.5, 2.8) [block, fill=black!15] {\tiny Encoder Dropout};
      \node (enc) at (-0.5, 2.2) [block, minimum width=2.0cm] {\tiny Sequence Encoder};
      \node (dec) at (0, 3.5) [block] {\tiny Dot Product};

      \draw [black!15, thick, fill=white] (1.05, 0.15) circle (0.15);
      \draw [black!15, thick, -latex] (1.05, 0.3) -- ($(emb.south)+(1.05, 0)$);
      \draw [black!30, thick, fill=white] (1.15, 0.15) circle (0.15);
      \draw [black!30, thick, -latex] (1.15, 0.3) -- ($(emb.south)+(1.15, 0)$);
      \foreach \x in {-1.25, -0.75, -0.25, 0.25, 1.25}{
        \draw [black, thick, fill=white] (\x, 0.15) circle (0.15);
        \draw [black, thick, -latex] (\x, 0.3) -- ($(emb.south)+(\x, 0)$);
      }
      \foreach \x in {-1.25, -0.75, -0.25, 0.25}{
        \draw [black, thick] ($(emb.north)+(\x, 0)$) -- ($(sdo.south)+(\x+0.5, 0)$);
        \draw [black, thick, -latex] ($(sdo.north)+(\x+0.5, 0)$) -- ($(enc.south)+(\x+0.5, 0)$);
      }
      \draw [black, thick] (enc.north) |- (edo.south);
      \draw [black, thick, -latex] (edo.north) -- ($(edo.north)+(0, 0.15)$) -- ($(edo.north)+(-0.75, 0.15)$) |- (dec.west);
      \draw [black, thick] ($(emb.north)+(1.25, 0)$) -- (cdo.south);
      \draw [black, thick] (cdo.north) -- (ffd.south);
      \draw [black, thick, -latex] (ffd.north) |- (dec.east);
    \end{tikzpicture}
    \label{fig-dnd}
  }
  \caption{Architecture of SBRSs}
  \label{fig-model}
\end{figure}
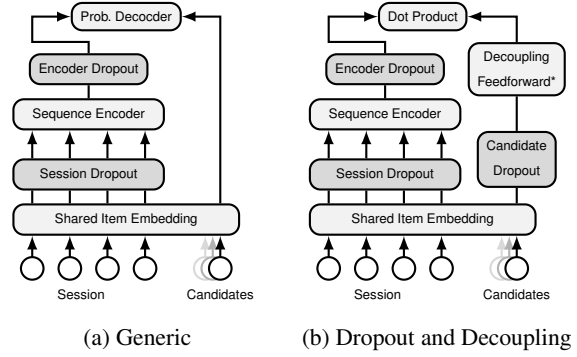

Before going into details, we will first briefly describe session-based recommender systems (SBRS). A SBRS takes a session, or any of its prefix sequence, as input, and outputs a probability distribution over all candidate interactions to be the next one in the sequence. A session is a bounded sequence of interactions of a user with the system. The boundary is often defined at the end of a long period with no user activity. Since SBRSs with mixed action types are not discussed here, we will use the term items, the objects users interact with, to replace interactions from here.

Figure \ref{fig-model} shows a high-level architecture of a general SBRS network. This architecture is adopted by the majority of state-of-the-art research \cite{gru4rec, narm, stamp, srgnn}. The input consists of two parts: a sequence of session items, $\left(x_{i_1}, ..., x_{i_T}\right)$, and a set of candidate items, $\left\{x_1, ..., x_N\right\}$. Both groups of items are converted into dense vectors by a shared item embedding. Then, the sequence of session item vectors, $\left(\mathbf{v}_{i_1}, ..., \mathbf{v}_{i_T}\right)$, is encoded into a single vector representation, $\mathbf{s}$, by a sequence encoder. Finally, both the session representation vector, $\mathbf{s}$, and the candidate item vectors, $\left\{\mathbf{v}_1, ..., \mathbf{v}_N\right\}$, are fed into a probability encoder to produce output probabilities, $\left\{(x_1, p_1), ..., (x_N, p_N)\right\}$.

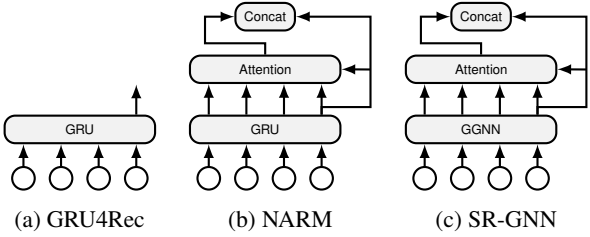
\begin{figure}[t]
  \centering
  \subfloat[GRU4Rec] {
    \begin{tikzpicture}
      \node (gru) at (0, 1.0) [block, minimum width=2.0cm] {\tiny GRU};
      \foreach \x in {-0.75, -0.25, 0.25, 0.75}{
        \draw [black, thick, fill=white] (\x, 0.35) circle (0.15);
        \draw [black, thick, -latex] (\x, 0.5) -- ($(gru.south)+(\x, 0)$);
      }
      \draw [black, thick, -latex] ($(gru.north)+(0.75, 0)$) -- ($(gru.north)+(0.75, 0.4)$);
    \end{tikzpicture}
    \label{fig-gru4rec}
  }
  \enspace
  \subfloat[NARM] {
    \begin{tikzpicture}
      \node (gru) at (0, 1.0) [block, minimum width=2.0cm] {\tiny GRU};
      \node (att) at (0, 1.8) [block, minimum width=2.0cm] {\tiny Attention};
      \node (cat) at (0, 2.5) [block] {\tiny Concat};
      \foreach \x in {-0.75, -0.25, 0.25, 0.75}{
        \draw [black, thick, fill=white] (\x, 0.35) circle (0.15);
        \draw [black, thick, -latex] (\x, 0.5) -- ($(gru.south)+(\x, 0)$);
        \draw [black, thick, -latex] ($(gru.north)+(\x, 0)$) -- ($(att.south)+(\x, 0)$);
      }
      \draw [black, thick, -latex] ($(gru.north)+(0.75, 0)$) |- ($(gru.north)+(1.4, 0.1)$) |- (cat.east);
      \draw [black, thick, -latex] ($(gru.north)+(1.4, 0.6)$) -- (att.east);
      \draw [black, thick, -latex] (att.north) |- ($(att.north)+(-0.8, 0.15)$) |- (cat.west);
    \end{tikzpicture}
    \label{fig-narm}
  }
  \enspace
  \subfloat[SR-GNN] {
    \begin{tikzpicture}
      \node (gru) at (0, 1.0) [block, minimum width=2.0cm] {\tiny GGNN};
      \node (att) at (0, 1.8) [block, minimum width=2.0cm] {\tiny Attention};
      \node (cat) at (0, 2.5) [block] {\tiny Concat};
      \foreach \x in {-0.75, -0.25, 0.25, 0.75}{
        \draw [black, thick, fill=white] (\x, 0.35) circle (0.15);
        \draw [black, thick, -latex] (\x, 0.5) -- ($(gru.south)+(\x, 0)$);
        \draw [black, thick, -latex] ($(gru.north)+(\x, 0)$) -- ($(att.south)+(\x, 0)$);
      }
      \draw [black, thick, -latex] ($(gru.north)+(0.75, 0)$) |- ($(gru.north)+(1.4, 0.1)$) |- (cat.east);
      \draw [black, thick, -latex] ($(gru.north)+(1.4, 0.6)$) -- (att.east);
      \draw [black, thick, -latex] (att.north) |- ($(att.north)+(-0.8, 0.15)$) |- (cat.west);
    \end{tikzpicture}
    \label{fig-srgnn}
  }
  \caption{Sequence encoders of SBRSs}
  \label{fig-encoder}
\end{figure}

In section \ref{sec-exp}, we will analyze Dropout and Decoupling on three representative SBRS models, which are GRU4Rec, NARM, and SR-GNN. All these three models fit into the architecture in figure \ref{fig-model}, and adopt linear-Softmax layers as their probability decoders. They only differ in their sequence encoders, which are shown in more detail in figure \ref{fig-encoder}. Using them as examples, we will show that Dropout and Decoupling could improve on a wide range of SBRS models.

\textbf{GRU4Rec} employs a GRU layer to encode session representations as $\mathbf{h}_T$ shown in equation \ref{eq-gru4rec}.
\begin{equation}
  \left(\mathbf{h}_i, ..., \mathbf{h}_T\right) = 
  \mathrm{GRU}\left(\mathbf{v}_{i_1}, ..., \mathbf{v}_{i_T}\right)
  \label{eq-gru4rec}
\end{equation}

\textbf{NARM} applies a dot-product attention layer to the GRU outputs. Details of the attention layer is shown in equations \ref{eq-narm}, where $q, c\in\mathbb{R}^D$ and $W_1, W_2\in\mathbb{R}^{D\times D}$ are model parameters. The bilinear layer described in the original paper is moved into probability decoders to fit in figure \ref{fig-model}, and the final session representation becomes $\mathbf{s}$.
\begin{equation}
  \begin{aligned}
    \alpha_t &= \mathbf{q}^\top \sigma\left(\mathbf{W}_1 \mathbf{h}_T + \mathbf{W}_2 \mathbf{h}_t + \mathbf{c} \right) \\
    \mathbf{h}_{att} &= \sum_{t=1}^T \alpha_t \mathbf{h}_t \\
    \mathbf{s} &= \left[\mathbf{h}_T; \mathbf{h}_{att}\right] \\
  \end{aligned}
  \label{eq-narm}
\end{equation}

\textbf{SR-GNN} replaces the GRU layer with a gated graph neural network (GGNN) \cite{ggnn}. It transforms session sequences into unweighted directed graphs. Item features are extracted by propagating information along both directions of the graph edges. The item features are then fed into the same dot-product attention layer in equations \ref{eq-narm}.

\begin{figure*}[t]
  \centering
  \subfloat[Linear] {
    \begin{tikzpicture}
      \node (hid) at (-1.0, 0.6) [block, minimum width=1.6cm] {\tiny Session Rep.};
      \node (can) at (1.0, 0.6) [block, minimum width=1.6cm] {\tiny Candidate Emb.};
      \node (lin) at (-1.0, 1.2) [block, minimum width=1.6cm] {\tiny Linear*};
      \node (dot) at (0, 1.8) [block] {\tiny Dot Product};
      \node (smx) at (0, 2.6) [block] {\tiny Softmax};
      \draw [black, thick] (hid.north) -- (lin.south);
      \draw [black, thick, -latex] (lin.north) |- (dot.west);
      \draw [black, thick, -latex] (can.north) |- (dot.east);
      \draw [black, thick, -latex] (dot.north) -- (smx.south);
    \end{tikzpicture}
    \label{fig-linear}
  }
  \quad
  \subfloat[Decoupled] {
    \begin{tikzpicture}
      \node (hid) at (-1.0, 0.6) [block, minimum width=1.6cm] {\tiny Session Rep.};
      \node (can) at (1.0, 0.6) [block, minimum width=1.6cm] {\tiny Candidate Emb.};
      \node (lin) at (-1.0, 1.2) [block, minimum width=1.6cm] {\tiny Linear*};
      \node (ffn) at (1.0, 1.2) [block, minimum width=1.6cm] {\tiny Feedforward};
      \node (dot) at (0, 1.8) [block] {\tiny Dot Product};
      \node (smx) at (0, 2.6) [block] {\tiny Softmax};
      \draw [black, thick] (hid.north) -- (lin.south);
      \draw [black, thick] (can.north) -- (ffn.south);
      \draw [black, thick, -latex] (lin.north) |- (dot.west);
      \draw [black, thick, -latex] (ffn.north) |- (dot.east);
      \draw [black, thick, -latex] (dot.north) -- (smx.south);
    \end{tikzpicture}
    \label{fig-decoupled}
  }
  \quad
  \subfloat[MLP] {
    \begin{tikzpicture}
      \node (hid) at (-1.0, 0.6) [block, minimum width=1.6cm] {\tiny Session Rep.};
      \node (can) at (1.0, 0.6) [block, minimum width=1.6cm] {\tiny Candidate Emb.};
      \node (lin1) at (-1.0, 1.2) [block, minimum width=1.6cm] {\tiny Linear};
      \node (lin2) at (1.0, 1.2) [block, minimum width=1.6cm] {\tiny Linear};
      \node (cat) at (0, 1.8) [block] {\tiny Entrywise Product};
      \node (mlp) at (0, 2.4) [block, minimum width=1.6cm] {\tiny MLP};
      \node (smx) at (0, 3.2) [block] {\tiny Softmax};
      \draw [black, thick] (hid.north) -- (lin1.south);
      \draw [black, thick] (can.north) -- (lin2.south);
      \draw [black, thick, -latex] (lin1.north) -- ($(lin1.north)+(0,0.1)$) -- ($(lin1.north)+(-0.4,0.1)$) |- (cat.west);
      \draw [black, thick, -latex] (lin2.north) -- ($(lin2.north)+(0,0.1)$) -- ($(lin2.north)+(0.4,0.1)$) |- (cat.east);
      \draw [black, thick] (cat.north) -- (mlp.south);
      \draw [black, thick, -latex] (mlp.north) -- (smx.south);
    \end{tikzpicture}
    \label{fig-mlp}
  }
  \quad
  \subfloat[MoS] {
    \begin{tikzpicture}
      \node (hid) at (-1.0, 0.6) [block, minimum width=1.6cm] {\tiny Session Rep.};
      \node (can) at (1.0, 0.6) [block, minimum width=1.6cm] {\tiny Candidate Emb.};
      \node (pool) at (0, 3.6) [block, minimum width=1.2cm] {\tiny Pooling};

      \node (lin) at (-1.2, 2.4) [block, minimum width=1.2cm] {\tiny Linear};
      \node (smx) at (-1.2, 3.0) [block] {\tiny Softmax};
      \draw [black, thick, -latex] (hid.north) -- 
        ($(hid.north)+(0,0.1)$) -- ($(hid.north)+(-1.0,0.1)$) |- 
        ($(lin.south)+(0, -0.3)$) -- (lin.south);
      \draw [black, thick] (lin.north) -- (smx.south);
      \draw [black, thick, -latex] (smx.north) |- (pool.west);
      \node at (-1.7, 3.5) [label] {\tiny Weights};

      \node (lin1) at (-0.8, 1.4) [black!15, block, minimum width=1.6cm] {\tiny Feedforward};
      \node (dot1) at (0.2, 2.0) [black!15, block] {\tiny Dot Product};
      \node (smx1) at (0.2, 2.8) [black!15, block] {\tiny Softmax};
      \draw [black!15, thick] ($(hid.north)+(0.2, 0)$) -- (lin1.south);
      \draw [black!15, thick, -latex] (lin1.north) |- (dot1.west);
      \draw [black!15, thick, -latex] ($(can.north)+(0.2, 0)$) |- (dot1.east);
      \draw [black!15, thick, -latex] (dot1.north) -- (smx1.south);
      \draw [black!15, thick, -latex] (smx1.north) -- ($(pool.south)+(0.2, 0)$);

      \node (lin2) at (-0.9, 1.3) [black!30, block, minimum width=1.6cm] {\tiny Feedforward};
      \node (dot2) at (0.1, 1.9) [black!30, block] {\tiny Dot Product};
      \node (smx2) at (0.1, 2.7) [black!30, block] {\tiny Softmax};
      \draw [black!30, thick] ($(hid.north)+(0.1, 0)$) -- (lin2.south);
      \draw [black!30, thick, -latex] (lin2.north) |- (dot2.west);
      \draw [black!30, thick, -latex] ($(can.north)+(0.1, 0)$) |- (dot2.east);
      \draw [black!30, thick, -latex] (dot2.north) -- (smx2.south);
      \draw [black!30, thick, -latex] (smx2.north) -- ($(pool.south)+(0.1, 0)$);

      \node (lin3) at (-1.0, 1.2) [block, minimum width=1.6cm] {\tiny Feedforward};
      \node (dot3) at (0, 1.8) [block] {\tiny Dot Product};
      \node (smx3) at (0, 2.6) [block] {\tiny Softmax};
      \draw [black, thick] (hid.north) -- (lin3.south);
      \draw [black, thick, -latex] (lin3.north) |- (dot3.west);
      \draw [black, thick, -latex] (can.north) |- (dot3.east);
      \draw [black, thick, -latex] (dot3.north) -- (smx3.south);
      \draw [black, thick, -latex] (smx3.north) -- (pool.south);
    \end{tikzpicture}
    \label{fig-mos}
  }
  \caption{Probability decoders of SBRSs}
  \label{fig-decoder}
\end{figure*}
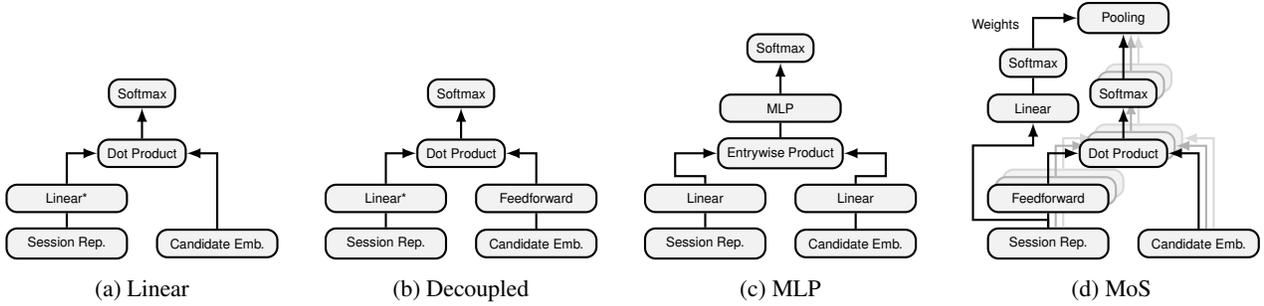

\subsection{Overfitting in Linear-Softmax Layers}

Overfitting is a phenomenon often associated with the model being more expressive than the data. However, it could still occur in networks as simple as a linear-Softmax layer. This counter-intuitive idea leaves this problem overlooked by previous work. Previous SBRS models mainly focus on the session and encoder dropout layers in figure \ref{fig-lin}. We propose to add another candidate dropout layer, shown in figure \ref{fig-dnd}. In section \ref{sec-exp}, we will show that it significantly improves the accuracy of a wide range of SBRS models on a variety of datasets.

A linear-Softmax-based SBRS could be described mathematically by equations \ref{eq-smxbasic}, where $\mathbf{v}_i \in \mathbb{R}^E$ is an item embedding, $f(\cdot)$ is the sequence encoder, and $\mathbf{s} \in \mathbb{R}^D$ is the session representation. $\mathbf{W} \in \mathbb{R}^{D\times E}$ is an optional matrix which is only required when the numbers of dimensions $D$ and $E$ are different.
\begin{equation}
  \begin{aligned}
    \mathbf{s} &= f\left(\mathbf{v}_{i_1}, ..., \mathbf{v}_{i_T}\right) \\
    \mathbf{p} &= \mathrm{softmax}\left(\mathbf{s}^\top \mathbf{W}[\mathbf{v}_1, ..., \mathbf{v}_N]\right) \\
  \end{aligned}
  \label{eq-smxbasic}
\end{equation}

During training, dropout layers are applied to the model. Equations \ref{eq-smxdropout} show this mathematically, and the last dropout function around $\mathbf{v}_i$ is the candidate dropout layer that we've been proposing.
\begin{equation}
  \begin{aligned}
    \mathbf{s} &= f\left(
      \mathrm{dropout}(\mathbf{v}_{i_1}), ..., 
      \mathrm{dropout}(\mathbf{v}_{i_T})\right) \\
    \mathbf{p} &= \mathrm{softmax}\left(
    \mathrm{dropout}(\mathbf{s})^\top \mathbf{W}\left[
    \mathrm{dropout}(\mathbf{v}_i)\right]_{i=1}^N\right) \\
  \end{aligned}
  \label{eq-smxdropout}
\end{equation}

\subsection{Tight-Coupling of Item Embedding Layers}

Sharing the same item embedding layer among session and candidate items creates a tight-coupling between sequence encoders and probability decoders. It's a common technique that's been widely used in linear-Softmax-based SBRSs, shown in figure \ref{fig-lin}. \cite{gru4rec, narm, srgnn}. In practice, it's crucial to share learned information between sequence encoders and probability decoders through item embeddings for models to reach state-of-the-art accuracy.

This tight-coupling could also stop accuracy from further improvement through interference. The shared item embeddings are simultaneously expected to be the input of sequence encoders and also the weights used to measure its similarity with session representations. Figure \ref{fig-linear} shows a block diagram of linear probability decoder. We propose to add a single feedforward layer, which consists of a linear layer with an activation function, in front of candidate embeddings, shown in figure \ref{fig-decoupled}. By introducing flexibility to the probability decoder, we will show that it alleviates the tight-coupling and improves accuracy in section \ref{sec-exp}.

\paragraph{Dropout and Decoupling} combines the two techniques mentioned above, shown in figure \ref{fig-dnd}. A candidate dropout layer is introduced to alleviate overfitting in the final linear layer. Depending on the dataset, an optional decoupling feedforward could be employed to loosen the tight-coupling between the sequence encoder and the probability decoder. We found Softplus works best among tanh, sigmoid, and ReLU as the activation function for the feedforward layer.

\subsection{Rank Deficiency}

Rank deficiency has been identified \cite{softmax} as the source of the Softmax bottleneck. The logits produced by a final linear layer form a low-rank matrix. Its rank is restrained by the number of hidden dimensions. MoS was proposed to tackle this issue. Figure \ref{fig-mos} shows a block diagram of MoS. The session representation is encoded into $K$ Softmax components by $K$ feedforward layers with $\tanh$ as activation functions. Then, the output probability distribution is calculated as a weighted average over the components.

In user-based recommender systems, many works have proposed different ways to introduce flexibility to the network, and reach higher accuracy than the traditional matrix factorization (MF) method, which includes the equivalent of a Linear-Softmax layer. In particular, the idea of combining MLP and factorization machine (FM) has inspired numerous research works. We will include a version of these networks as probability decoder in section \ref{sec-exp}, and compare it against MoS and our D\&D.

\paragraph{MLP} model is shown by a block diagram in figure \ref{fig-mlp}. Inspired by the idea of combining MLP and FM, we stack a 2-layer MLP on top of a FM, which consists of an entrywise product of a session representation and a candidate item embedding both transformed by an individual linear layer.

\subsection{Time Complexity Analysis}

\begin{table}[t]
  \centering
  \small
  \begin{tabular}{lll}
  \toprule
  Decoder & Training & Query \\
  \midrule
  Linear & $\mathcal{O}(NMD)$ & $\mathcal{O}(D\ln{M} + L\ln{L})$  \\
  Decoupled & $\mathcal{O}(NMD+MD^2)$ & $\mathcal{O}(D\ln{M} + L\ln{L})$ \\
  MLP & $\mathcal{O}(NMD^2)$ & $\mathcal{O}(MD^2)$ \\
  MoS & $\mathcal{O}(NMDK)$ & $\mathcal{O}(MDK)$ \\
  \bottomrule
  \end{tabular}
  \caption{Time complexity of probability decoders ($N$ is the batch size. $M$ is the candidate pool size. D is the number of hidden dimensions. $L$ is the size limit of query results. $K$ is the number of MoS components.)}
  \label{tbl-time}
\end{table}

Time complexity is one important aspect of SBRSs during both training and query. Introducing complexity to the network could result in dramatic change in time complexity. During query, the time complexity is even more constrained. Breaking existing acceleration methods or requiring special operations could result in significant costs. We will focus our analysis on probability decoders in figure \ref{fig-decoder}.

Table \ref{tbl-time} summarizes the time complexity of different probability decoders. We assume that the candidate pool size $M$ is significantly larger than the other variables, which is over $10^4$ in all our datasets. The batch number $N$, the number of hidden dimensions $D$, the size limit of query results $L$ are of similar magnitude, which are usually around $10^2$. The number of MoS components $K$ is usually  around $10^1$.

\paragraph{Training Time Complexity}
The time complexity of the linear decoder is $\mathcal{O}(NMD)$. The optional linear layer is omitted from the formula since the sequence encoder must have greater or equal terms. The decoupled decoder adds a feedforward layer to the candidate item embeddings. Since embeddings are the same for the whole batch, it could be calculated only once in each training step. Therefore, there's only an additional term $\mathcal{O}(MD^2)$, which could be omitted when $M$ and $N$ are of similar magnitude. The MLP decoder will increase the time complexity significantly to $\mathcal{O}(NMD^2)$. It might require candidate selection mechanisms to control the size of $M$ and make training efficient. The MoS decoder has $K$ Softmax components, and its time complexity is also $K$ times higher.

\paragraph{Query Time Complexity}
The time complexity of the linear decoder is $\mathcal{O}(D\ln{M} + L\ln{L})$ for two reasons. First, since Softmax function preserves the order of logit values, it could be ignored for ranking. Second, a pre-built spatial index could be used to retrieve the top $L$ dot-product values in log time. The decoupled decoder has the same time complexity as the linear decoder, because the results of the feedforward layer could be pre-computed. The MLP and MoS decoder could not be accelerated by conventional spatial indices. Therefore, their time complexities are linear to the candidate pool size $M$. Furthermore, the MoS decoder will require extra operations to calculate the denominator of $K$ Softmax components, shown in equation \ref{eq-mosprob}. They both might need candidate selection mechanisms during  query, which could hurt overall accuracy.

\section{Experiments} \label{sec-exp}

In this section, we will first describe the setup of our experiments. Then, we will analyze the experiment results, and show the effect of Dropout and Decoupling.

\subsection{Datasets and Preprocessing}

We conducted our experiments on the following public real-world datasets from a wide range of different applications.

\begin{itemize}
  \item \textbf{Diginetica}\footnote{http://cikm2016.cs.iupui.edu/cikm-cup} was released at CIKM Cup 2016. It contains anonymized search and browsing logs collected from an e-commerce search engine over 6 months.
  \item \textbf{RetailRocket}\footnote{http://www.kaggle.com/retailrocket/ecommerce-dataset} was released in a Kaggle contest. It contains sessions collected from a real-world e-commerce website over 5 months.
  \item \textbf{Gowalla}\footnote{http://snap.stanford.edu/data/loc-gowalla.html} \cite{gowalla} contains check-in data collected from a geo-location-based social networking website over 21 months.
\end{itemize}

In preprocessing, we follow the common practice \cite{narm, stamp, srgnn} to remove items appearing less than 5 times and sessions with length 1. For Gowalla dataset, we split the user timelines into disjoint sessions at the end of each inactive period longer than 8 hours, and only keep items appearing no less than 20 times. Table \ref{tlb-dataset} shows some statistics of the datasets after preprocessing, where \# sessions does not include prefix sequences. In each session sequence, every element and its non-empty preceding sequence will be used as target and input of an individual sample. Preceding sequences are truncated to the latest 20 elements if its length exceeds this limit.

\begin{table}[t]
  \centering
  \small
  \begin{tabular}{crrrrr}
    \toprule
    Dataset &
    \multicolumn{1}{c}{\# events} & 
    \multicolumn{1}{c}{\# sessions} & 
    \multicolumn{1}{c}{\# items} & 
    avg. len.\\
    \midrule
    Diginetica & 993,483 & 204,789 & 43,136 & 5.13 \\
    RetailRocket & 1,188,938 & 323,002 & 50,449 & 7.56 \\
    Gowalla & 1,195,940 & 371,191 & 30,774 & 5.11 \\
    \bottomrule
  \end{tabular}
  \caption{Statistics of datasets}
  \label{tlb-dataset}
\end{table}

\subsection{Algorithms}

We applied our method to sequence decoders of the following representative works in SBRS to show that it's a general solution to the problems we mentioned. The encoders are also shown in detail in figure \ref{fig-encoder}.

\begin{itemize}
  \item \textbf{GRU4Rec} \cite{gru4rec} is a RNN-based model, which encodes session representations using GRU layers.
  \item \textbf{NARM} \cite{narm} employs RNNs with attention to capture users' main purposes and sequential behaviors.
  \item \textbf{SR-GNN} \cite{srgnn} employs GGNNs with attention to obtain accurate item features through taking complex transitions of items into account.
\end{itemize}

We compared our decoupled decoder to the following decoders, which are shown in detail in figure \ref{fig-decoder}.

\begin{itemize}
  \item \textbf{Linear} consists of a final dot-product layer.
  \item \textbf{Decoupled} passes candidate item embeddings through a feedforward layer before the final dot-product layer.
  \item \textbf{MLP} performs entrywise products between candidate embeddings and session representations, and then passes the result through a final MLP layer.
  \item \textbf{MoS} \cite{softmax} recommends items similar to the previous item in the current session. Similarity is measured by a cosine index on the co-occurrence of items in the same session.
\end{itemize}

We also included some baseline algorithms for comparison, which could reveal different properties of each dataset.

\begin{itemize}
  \item \textbf{POP} recommends the most frequent items in training set.
  \item \textbf{S-POP} recommends the most frequent items in the current session.
  \item \textbf{Item-KNN} recommends items similar to the previous item in the session. Similarity is measured by a cosine index on co-occurence of items in the same session.
\end{itemize}

\subsection{Evaludation Protocol}

We adopted the commonly used top-k metrics \cite{gru4rec, narm, stamp, srgnn} in our evaluation, which are HR@20 (hit rate, also called P@20 and Recall@20 with equivalent definitions) and MRR@20 (mean reciprocal rank). In order to produce unbiased results, each dataset is split into training, validation, and test sets. The most recent 10\% of the time period is reserved for the test set, and 10\% of the rest sessions are reserved for the validation set. Reported metrics are measured on the test set when the same metric reaches the highest point on the validation set. This prevents information from test and validation sets to leak into the model. Each datapoint is an average of 5 individual runs.

\subsection{Parameter Setup}

In order to have a common ground for comparison, the dimension of all hidden states and item embeddings are set to 100 for all models. We followed \cite{gru4recp, narm} to set the dropout ratios of GRU4Rec and NARM to \nicefrac{1}{4} for session embeddings and 0 and \nicefrac{1}{2} for encoder outputs, respectively. We adopted the same dropout scheme for SR-GNN as NARM, because we found it produces more stable results. In training, Adam is employed for optimization. The batch size is 200 and the learning rate is $10^{-3}$. There's no learning rate schedule and weight decay. For hyper-parameter tuning, we explored the following ranges: \{0, \nicefrac{1}{8}, \nicefrac{1}{4}, \nicefrac{3}{8}, \nicefrac{1}{2}\} for candidate dropout ratio, \{1, 2, 3\} for number of MLP layers, \{2, 3, 4, 6, 8\} for number of MoS components.

\subsection{Performance Comparison}

\begin{table}[t]
  \centering
  \small
  \setlength\tabcolsep{5pt}
  \begin{tabular}{ccccccc}
    \toprule
    \multirow{4}{*}{Model} & 
    \multicolumn{2}{c}{Diginetica} & 
    \multicolumn{2}{c}{RetailRocket} & 
    \multicolumn{2}{c}{Gowalla} \\
    \cmidrule(lr){2-3}\cmidrule(lr){4-5}\cmidrule(lr){6-7}
    & HR & MRR & HR & MRR & HR & MRR \\
    \cmidrule(lr){2-2}\cmidrule(lr){3-3}\cmidrule(lr){4-4}
    \cmidrule(lr){5-5}\cmidrule(lr){6-6}\cmidrule(lr){7-7}
    & @20 & @20 & @20 & @20 & @20 & @20 \\
    \midrule
    \multicolumn{7}{c}{Baseline} \\
    \midrule
    POP      & \enspace 0.84 & \enspace 0.19 & \enspace 1.61 & \enspace 0.51 & \enspace 3.78 & \enspace 0.75 \\
    S-POP    & 20.28 & 10.96 & 41.91 & 31.57 & \enspace 9.31 & \enspace 5.33 \\
    Item-KNN & 30.44 & \enspace 9.20 & 18.35 & \enspace 6.92 & 39.37 & 17.36 \\
    \midrule
    \multicolumn{7}{c}{Decoder / Linear} \\
    \midrule
    GRU4Rec  & 48.56 & 16.77 & 58.58 & 37.01 & 55.12 & \U{25.88} \\
    NARM     & 50.72 & 17.44 & 61.38 & 37.49 & 56.81 & 25.46 \\
    SR-GNN   & 50.53 & 17.45 & \U{62.27} & 38.45 & 56.61 & 25.30 \\
    \midrule
    \multicolumn{7}{c}{Decoder / D\&D} \\
    \midrule
    GRU4Rec  & \B{51.79} & \B{18.34} & 62.24 & \U{39.04} & \B{58.48} & \B{26.68} \\
    NARM     & \U{51.75} & \U{18.13} & 62.20 & 38.60 & \U{57.77} & 25.65 \\
    SR-GNN   & 51.52 & 18.01 & \B{62.77} & \B{39.28} & 57.43 & 25.80 \\
    \bottomrule
  \end{tabular}
  \caption{Baseline Comparison}
  \label{tlb-model}
\end{table}

\begin{table}[t]
  \centering
  \small
  \setlength\tabcolsep{5pt}
  \begin{tabular}{ccccccc}
    \toprule
    \multirow{4}{*}{Decoder} & 
    \multicolumn{2}{c}{Diginetica} & 
    \multicolumn{2}{c}{RetailRocket} & 
    \multicolumn{2}{c}{Gowalla} \\
    \cmidrule(lr){2-3}\cmidrule(lr){4-5}\cmidrule(lr){6-7}
    & HR & MRR & HR & MRR & HR & MRR \\
    \cmidrule(lr){2-2}\cmidrule(lr){3-3}\cmidrule(lr){4-4}
    \cmidrule(lr){5-5}\cmidrule(lr){6-6}\cmidrule(lr){7-7}
    & @20 & @20 & @20 & @20 & @20 & @20 \\
    \midrule
    \multicolumn{7}{c}{Encoder / GRU4Rec} \\
    \midrule
    Linear-$\delta$    & 51.78 & 18.19 & 62.26 & 38.38 & 58.45 & \U{26.67} \\
    Decoulped-$\delta$ & 51.40 & \U{18.33} & 62.03 & 39.01 & 57.50 & 26.34 \\
    MLP-$\delta$       & \U{51.96} & 18.26 & 62.60 & 38.80 & 58.57 & 23.49 \\
    MoS-$\delta$       & 51.21 & 18.27 & 61.87 & 38.87 & 57.58 & \B{26.71} \\
    \midrule
    \multicolumn{7}{c}{Encoder / NARM} \\
    \midrule
    Linear-$\delta$    & 51.18 & 17.60 & 62.00 & 37.77 & 57.41 & 25.64 \\
    Decoulped-$\delta$ & 51.76 & 18.13 & 62.16 & 38.61 & 57.79 & 25.57 \\
    MLP-$\delta$       & \B{52.11} & 17.75 & \B{63.09} & 38.42 & \B{58.84} & 23.06 \\
    MoS-$\delta$       & 51.31 & 17.90 & 62.08 & 39.06 & 57.90 & 25.81 \\
    \midrule
    \multicolumn{7}{c}{Encoder / SR-GNN} \\
    \midrule
    Linear-$\delta$    & 51.33 & 17.78 & 62.79 & 38.68 & 57.17 & 25.79 \\
    Decoulped-$\delta$ & 51.54 & 18.03 & 62.76 & \U{39.26} & 57.43 & 25.30 \\
    MLP-$\delta$       & 51.77 & 17.71 & \U{62.93} & 38.60 & \U{58.67} & 23.31 \\
    MoS-$\delta$       & 51.55 & \B{18.39} & 62.49 & \B{39.70} & 57.79 & 26.07 \\
    \bottomrule
  \end{tabular}
  \caption{Decoder Comparison}
  \label{tlb-encoder}
\end{table}

Intended to reveal basic properties of each dataset, the comparison of baseline models is shown in the top part of table \ref{tlb-model}. POP shows popularity-based recommendations generally perform poorly, while it's slightly better in Gowalla than in the others. S-POP shows session sequences are the most repetitive in RetailRocket and the least in Gowalla. Item-KNN shows the co-occurrence relation between items are the strongest in Gowalla and the weakest in RetailRocket.
With the introduction of the attention mechanism and graph neural network, NARM and SR-GNN notably improve the performance over the pure RNN-based GRU4Rec on all 3 datasets. The comparison of linear-Softmax models is shown in the middle part of table \ref{tlb-model}.

Dropout and Decoupling significantly improve the performance of all sequence encoders on all datasets. Details are shown in the bottom part of table \ref{tlb-model}. There's two main effects to observe.
First, the differences between sequence encoders become less significant. While our method enables all 3 sequence encoders to reach state-of-the-art performance level, the gap between them shrinks. This implies the overfitting and tight-coupling problems account for a significant part of the performance gap.
Second, our method improves the most on RNN sequence encoder. It allows GRU4Rec to outperform more complex algorithms, NARM and SR-GNN, on Diginetica and Gowalla. It implies GRU4Rec is more susceptible to overfitting and tight-coupling. Our method provides an opportunity to revitalize RNN-based models.

The detailed comparison of probability decoders is shown in table \ref{tlb-encoder}, where the suffix $\delta$ indicates there's a candidate dropout layer. 
First, let's have a more detailed discussion on the optional decoupling feedforward layer in our method by comparing Linear-$\delta$ and Decoupled-$\delta$. While overfitting is a general problem in linear-Softmax models, tight-coupling does not always harm performance. Decoupling generally improves performance with exceptions for GRU4Rec encoder and Gowalla dataset, where its effect is mixed.
Second, more complex decoders, MLP-$\delta$ and MoS-$\delta$, could reach even higher performance metrics with much greater computation costs. There's still no decoder that outperforms or underperforms others in all situations.

\subsection{Ablation Study}

\begin{table}[t]
  \centering
  \small
  \setlength\tabcolsep{5pt}
  \begin{tabular}{ccccccc}
    \toprule
    \multirow{4}{*}{Decoder} & 
    \multicolumn{2}{c}{Diginetica} & 
    \multicolumn{2}{c}{RetailRocket} & 
    \multicolumn{2}{c}{Gowalla} \\
    \cmidrule(lr){2-3}\cmidrule(lr){4-5}\cmidrule(lr){6-7}
    & HR & MRR & HR & MRR & HR & MRR \\
    \cmidrule(lr){2-2}\cmidrule(lr){3-3}\cmidrule(lr){4-4}
    \cmidrule(lr){5-5}\cmidrule(lr){6-6}\cmidrule(lr){7-7}
    & @20 & @20 & @20 & @20 & @20 & @20 \\
    \midrule
    \multicolumn{7}{c}{Encoder / GRU4Rec} \\
    \midrule
    SepEmb    & 37.93 & 12.05 & 49.01 & 28.96 & 48.27 & 23.84 \\
    Linear    & 48.56 & 16.77 & 58.58 & 37.01 & 55.12 & 25.88 \\
    Decoulped & 48.98 & 17.38 & 59.17 & 37.63 & 54.97 & 25.67 \\
    MLP       & 49.58 & 17.86 & 60.82 & 38.70 & 56.09 & 23.31 \\
    MoS       & 48.50 & 16.87 & 59.37 & 38.10 & 54.62 & 25.61 \\
    \midrule
    \multicolumn{7}{c}{Encoder / NARM} \\
    \midrule
    SepEmb    & 41.52 & 12.93 & 52.37 & 30.02 & 51.35 & 24.14 \\
    Linear    & 50.72 & 17.44 & 61.38 & 37.49 & 56.81 & 25.46 \\
    Decoulped & 51.27 & 17.79 & 61.79 & 37.94 & 56.92 & 25.23 \\
    MLP       & 51.34 & 17.66 & 62.68 & 38.42 & 58.01 & 23.06 \\
    MoS       & 50.91 & 17.49 & 61.80 & 38.69 & 57.35 & 25.67 \\
    \midrule
    \multicolumn{7}{c}{Encoder / SR-GNN} \\
    \midrule
    SepEmb    & 40.59 & 12.40 & 52.77 & 30.31 & 50.83 & 23.94 \\
    Linear    & 50.53 & 17.45 & 62.27 & 38.45 & 56.61 & 25.30 \\
    Decoulped & 50.81 & 17.61 & 62.14 & 38.45 & 56.85 & 25.27 \\
    MLP       & 51.35 & 17.70 & 62.65 & 38.60 & 57.95 & 23.31 \\
    MoS       & 50.68 & 17.84 & 62.34 & 39.42 & 57.17 & 25.77 \\
    \bottomrule
  \end{tabular}
  \caption{Ablation Study}
  \label{tlb-dropout}
\end{table}

The importance of sharing item embeddings is shown by comparing SepEmb and Linear in table \ref{tlb-dropout}. SepEmb adopts separated embeddings for session and candidate items. The results show that metrics could drop significantly when sharing is not employed, especially in dataset Diginetica and RetailRocket. This partially explains why decoupling is more effective in these datasets, since the coupling tightens as the model learns more through shared embeddings.

The effect of candidate dropout layers is also shown by comparing table \ref{tlb-dropout} to table \ref{tlb-encoder}. Universally, candidate dropout layers improve the performance of all models on all datasets in our evaluation. Conventional wisdom often associates overfitting with complex networks. In our experiments, candidate dropout layers are shown to be more effective in simpler networks, such as GRU4Rec encoder, Linear and Decoupled decoders. This implies that these simpler networks are actually more prone to overfitting.

\section{Conclusions}

The Softmax bottleneck is a serious problem in session-based recommender systems, that hasn't been examined by previous work. In this paper, the problem is investigated and more aspects of this issue are introduced. Overfitting in linear-Softmax layers and tight-coupling of item embeddings are identified as two major contributors to the bottleneck. Dropout and Decoupling (D\&D), a simple yet effective method, is proposed to alleviate this bottleneck. Comprehensive
experiments confirm that the proposed method can improve on a wide range of state-of-the-art models, while keeping the same time complexity during training and query.

\bibliographystyle{unsrtnat}
\bibliography{dnd}

\end{document}